\documentclass[letterpaper, 10 pt, conference]{ieeeconf}  

\usepackage{graphicx} 
\usepackage{amsmath} 
\usepackage{array}
\usepackage{multirow} 
\usepackage{amsfonts}
\usepackage{float}
\usepackage{subfig} 
\usepackage{lipsum} 
\usepackage[
    colorlinks=true,    
    linkcolor=red,      
    citecolor=blue,     
    urlcolor=magenta    
]{hyperref}

\usepackage{caption}
\IEEEoverridecommandlockouts                              

\title{\LARGE \bf
Voxel Densification for Serialized 3D Object Detection: Mitigating Sparsity via Pre-serialization Expansion
}

\author{Qifeng Liu, Dawei Zhao, Yabo Dong, Linzhi Shang, Liang Xiao, Juan Wang, Kunlong Zhao,Dongming Lu, Qi Zhu}

\begin{document}

\maketitle
\thispagestyle{empty}
\pagestyle{empty}

\begin{abstract}
Recent advances in point cloud object detection have increasingly adopted Transformer-based and State Space Models (SSMs) to capture long-range dependencies. However, these serialized frameworks strictly maintain the consistency of input and output voxel dimensions, inherently lacking the capability for voxel expansion. This limitation hinders performance, as expanding the voxel set is known to significantly enhance detection accuracy, particularly for sparse foreground objects. To bridge this gap, we propose a novel Voxel Densification Module (VDM). Unlike standard convolutional stems, VDM is explicitly designed to promote \textit{pre-serialization spatial expansion}. It leverages sparse 3D convolutions to propagate foreground semantics to neighboring empty voxels, effectively densifying the feature representation before it is flattened into a sequence. Simultaneously, VDM incorporates residual sparse blocks to aggregate fine-grained local context, ensuring rich geometric feature extraction. To balance the computational overhead of increased voxel density, we introduce a strategic cascaded downsampling mechanism. We integrate VDM into both Transformer-based (DSVT) and SSM-based (LION) detectors. Extensive experiments demonstrate that VDM consistently improves detection accuracy across multiple benchmarks. Specifically, our method achieves \textbf{74.8 mAPH (L2)} on the Waymo \textit{validation set} and \textbf{70.5 mAP} on the nuScenes \textit{test set}. Furthermore, it attains \textbf{42.6 mAP} on the Argoverse 2 \textit{validation set} and \textbf{67.6 mAP} on the ONCE \textit{validation set}, consistently outperforming the baseline models. The source code will be made publicly available at \url{https://github.com/qifeng22/VDM}.
\end{abstract}


\section{Introduction}
3D object detection based on point clouds plays a crucial role in applications such as autonomous driving~\cite{bansal2018chauffeurnet} and robotic navigation~\cite{zhu2017target}. Inspired by the remarkable success of Transformers~\cite{vaswani2017attention} and State Space Models (SSMs)~\cite{gu2023mamba} in natural language processing, researchers have recently explored adapting these architectures to computer vision tasks, achieving strong performance. For example, in the domain of point cloud 3D object detection, DSVT~\cite{wang2023dsvt} introduces a dynamic sparse window attention mechanism that adaptively partitions each window into local sub-regions based on voxel sparsity, enabling fully parallel attention computations. LION~\cite{liu2024lion} further extends this direction by employing large grouping windows and leveraging SSMs' ability to capture long-range voxel dependencies, achieving impressive performance on several public datasets.

Despite these advancements, a fundamental limitation persists in current serialized frameworks: the \textit{spatial-serial gap}. Both Transformer-based and SSM-based models typically flatten sparse voxels into 1D sequences to model long-range dependencies. However, this serialization process strictly maintains the input-output token consistency, meaning the set of active voxels remains static throughout the backbone. Consequently, these models lack the inherent capability to perform \textit{spatial expansion} that is trivial in convolutional networks. This limitation is critical because raw point clouds are often sparse and incomplete; without expanding the voxel set to include neighboring empty regions, the model struggles to capture sufficient local context for accurate localization, particularly for distant or occluded objects. While prior works like LION~\cite{liu2024lion} attempt to mitigate this via intermediate feature diffusion, they operate on already-serialized features, where spatial topology is disrupted. In contrast, we argue that the optimal strategy is to explicitly densify the voxel representation \emph{before} serialization, thereby bridging the gap between spatial completeness and sequential modeling efficiency. 

\begin{figure}[tbp]
    \centering
    \includegraphics[width=0.95\linewidth]{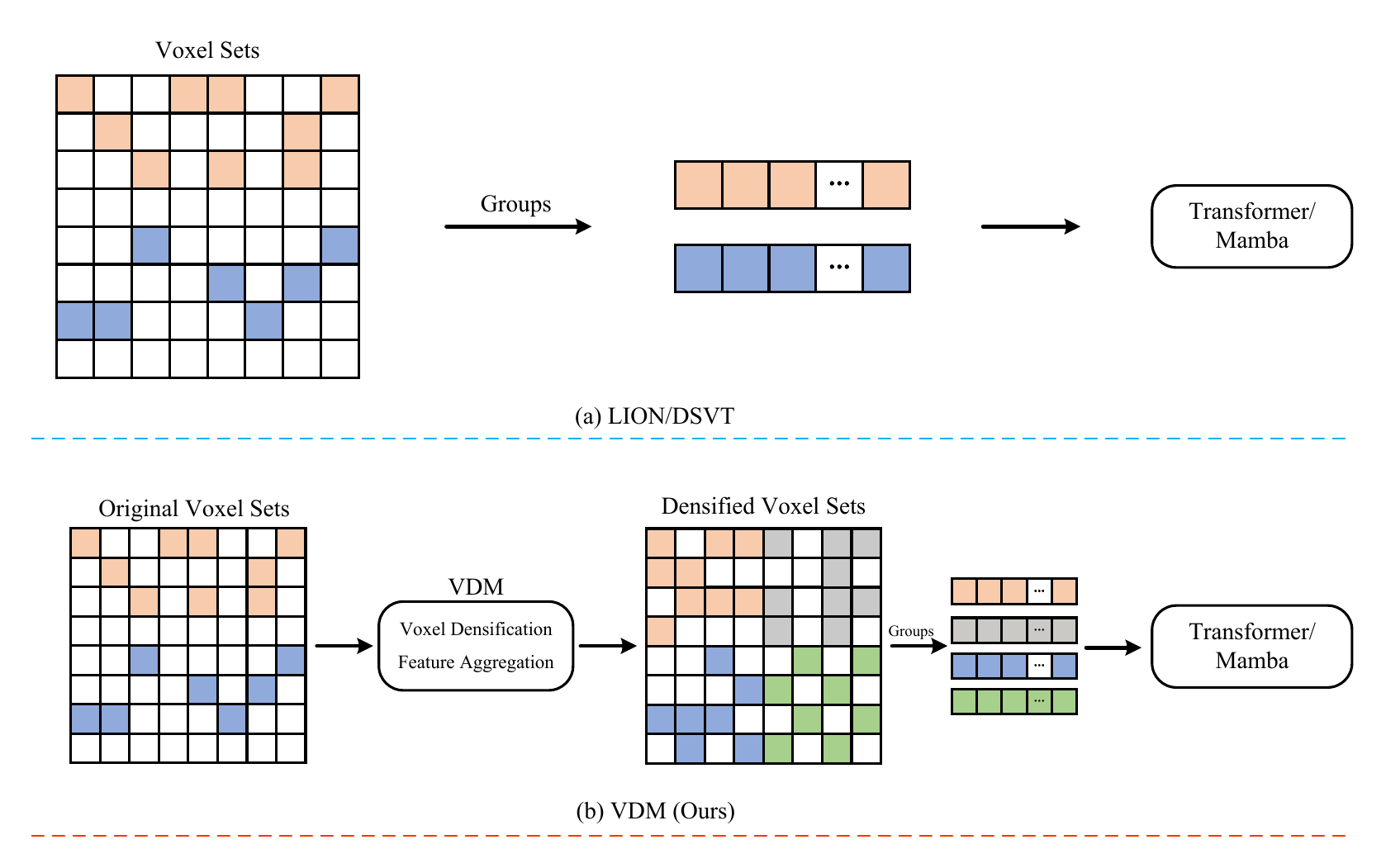}
    \caption{Comparison of serialized detection paradigms. (a) Existing methods such as LION~\cite{liu2024lion} and DSVT~\cite{wang2023dsvt} directly process sparse voxel sets. (b) Our proposed framework incorporates the Voxel Densification Module (VDM) to explicitly densify and aggregate voxel features before sequence modeling.}
    \label{fig:introduction}
\end{figure}

Motivated by the spatial flexibility of CNN-based architectures, we propose a novel \textbf{Voxel Densification Module (VDM)}, as illustrated in Figure~\ref{fig:introduction}, specifically tailored to serve as a pre-serialization stem for Mamba-based and Transformer-based 3D detectors. Positioned prior to the flattening of voxel sequences, VDM leverages sparse 3D convolutional operations to explicitly \textit{expand} the foreground voxel set, effectively densifying the feature representation. This process generates new informative voxels in the 3D neighborhood of the original sparse points, ensuring that the subsequent serialized model receives a spatially enriched input. Structurally, VDM integrates stacked sparse 3D convolutions with residual blocks to achieve two goals simultaneously: (1) \textbf{Voxel Densification}, which propagates semantics to empty spatial neighbors, and (2) \textbf{Fine-grained Aggregation}, which captures local geometric details. Crucially, to balance the computational cost of processing increased voxel counts, we introduce a strategic downsampling mechanism (reducing resolution to 1/4), optimizing the trade-off between spatial density and sequence length.

VDM is designed to be modular and general, allowing seamless integration with both Transformer-based and SSM-based detectors. To evaluate its effectiveness, we perform extensive experiments on four large-scale datasets. Our VDM-enhanced SSM-based frameworks achieve \textbf{74.8 mAPH (L2)} on the Waymo validation set, \textbf{70.5 mAP} on the nuScenes test set, \textbf{42.6 mAP} on the Argoverse 2 validation set, and \textbf{67.6 mAP} on the ONCE validation set, consistently outperforming baseline models across different benchmarks. Additionally, the Transformer-based variant (VDM-DSVT) improves performance by \textbf{+1.2 mAPH (L2)} over the DSVT baseline on the Waymo dataset, confirming the general effectiveness of our approach.

\noindent\textbf{Our contributions can be summarized as follows:}
\begin{itemize}
    \item We identify the limitation of static voxel sets in serialized 3D detectors and propose \textbf{VDM} (Voxel Densification Module), a plug-and-play operator that explicitly densifies sparse voxel inputs via pre-serialization convolution.
    \item We design a dual-path architecture within VDM that simultaneously performs spatial expansion (to increase foreground recall) and fine-grained feature aggregation (to enhance local context).
    \item VDM demonstrates universal compatibility, seamlessly integrating with both Transformer-based and SSM-based backbones to significantly boost performance.
    \item We validate the effectiveness of VDM through extensive experiments, achieving state-of-the-art results across Waymo, nuScenes, Argoverse 2, and ONCE datasets.
\end{itemize}

\section{Related Work}

3D object detection based on point clouds has seen significant progress. Existing approaches can be broadly categorized into CNN-based, Transformer-based, and State Space Model (SSM)-based methods. While CNN-based models have established a strong baseline, recent serialization-based architectures (Transformers and SSMs) have shown superior potential in capturing long-range dependencies, though they face distinct challenges regarding spatial continuity.

\noindent \textbf{CNN-Based 3D Object Detection.} CNN-based detectors are the pioneers in this field, utilizing either voxel-based or point-based representations. VoxelNet~\cite{zhou2018voxelnet} introduces Voxel Feature Encoding (VFE) to generate unified feature representations. To improve efficiency, PointPillars~\cite{lang2019pointpillars} collapses point clouds into vertical columns, enabling fast 2D convolution processing. CenterPoint~\cite{centerpoint} advances this by adopting a center-based anchor-free detection head. More recently, methods like HEDNet~\cite{hednet} utilize encoder-decoder designs for long-range perception. Notably, SAFDNet~\cite{SAFDnet} identifies the issue of feature sparsity and proposes an adaptive feature diffusion strategy. However, SAFDNet primarily focuses on diffusing features within the backbone for CNN-based detectors. In contrast, our work targets the specific "spatial-serial gap" in serialization-based models, aiming to physically densify the voxel set prior to sequence flattening.

\noindent \textbf{Transformer-Based 3D Object Detection.} Transformers have demonstrated state-of-the-art performance by treating voxels as sequences. PVT-SSD~\cite{pvtssd} employs a Point-Voxel Transformer to capture global dependencies. FlatFormer~\cite{liu2023flatformer} improves efficiency by flattening point clouds via window-based sorting. DSVT~\cite{wang2023dsvt} introduces Dynamic Sparse Window Attention to handle sparsity efficiently. Despite their success, these methods typically operate on a \textit{static} set of non-empty voxels. Due to the strict input-output consistency of attention mechanisms, they lack the inherent ability to expand the voxel set to neighboring empty regions, often leading to insufficient context for sparse objects.

\noindent \textbf{SSM-Based 3D Object Detection.} Mamba-based SSMs represent a burgeoning direction. LION~\cite{liu2024lion} leverages SSMs with large window grouping to model wider spatial contexts. Voxel Mamba~\cite{voxel-mamba} serializes the entire voxel space to capture global dependencies via Dual-Scale SSM Blocks. UniMamba~\cite{jin2025unimamba} integrates 3D convolutions with SSMs in a multi-head architecture. While UniMamba combines CNNs and SSMs, it treats them as parallel or interleaved feature extractors. 

Different from these approaches, our proposed \textbf{Voxel Densification Module (VDM)} is explicitly designed as a \textit{pre-serialization stem}. Instead of merely extracting features, VDM fundamentally alters the spatial distribution of the input by densifying foreground voxels. This ensures that the downstream serialized models (whether Transformer or SSM) receive a spatially enriched token sequence, directly addressing the sparsity limitation inherent in serialization.

\begin{figure*}[t]
    \centering
    \includegraphics[width=0.95\textwidth]{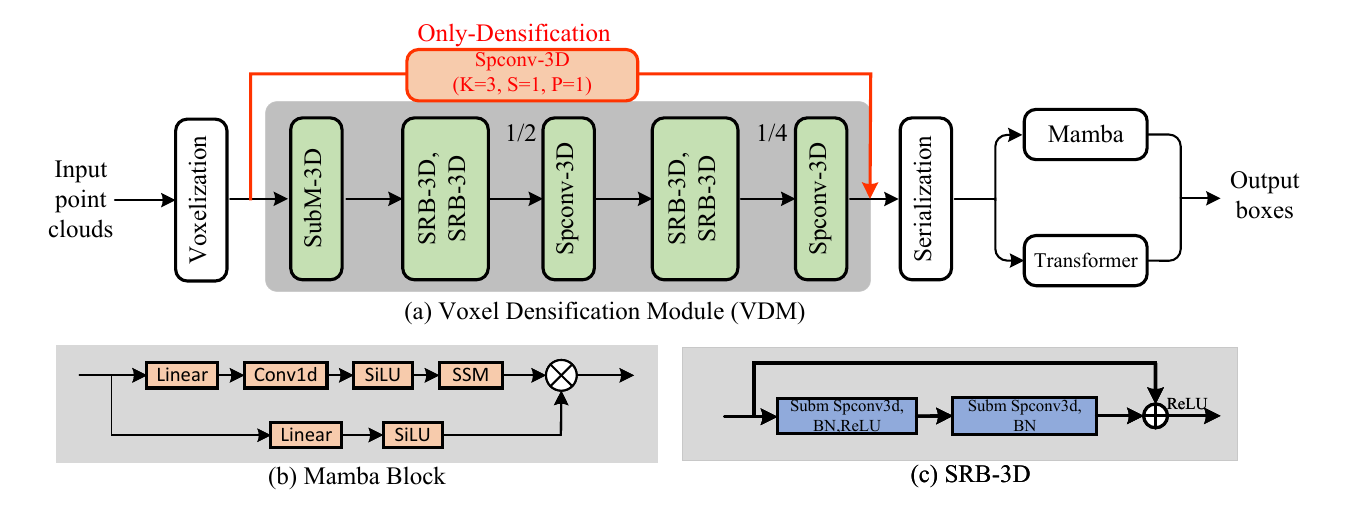}
    \caption{Overall architecture of our proposed method. (a) The \textbf{Voxel Densification Module (VDM)} is positioned prior to serialization, specifically designed to \textit{densify} sparse voxel inputs and aggregate local context. The core VDM path (gray box) utilizes stacked Submanifold 3D Convolutions (SubM-3D) and Sparse Residual Blocks (SRB-3D) with downsampling layers (Spconv-3D) to reduce resolution to 1/4. (b) The Mamba Block used in SSM-based backbones. (c) The detailed structure of the Sparse Residual Block (SRB-3D). \textbf{Note:} The red path labeled ``\textbf{Only-Densification}'' represents the ablation variant where only a single expansion layer (Spconv-3D with $K=3, S=1$) is applied, isolating the efficacy of the voxel expansion mechanism.}
    \label{fig:framework}
\end{figure*}

\section{Method}

In this section, we introduce the proposed \textbf{Voxel Densification Module (VDM)}, designed to enhance the representation of voxelized point cloud features. VDM leverages convolutional operations to perform early-stage \textit{voxel expansion}, increasing the density of meaningful voxel information. This provides richer input for downstream serialized models such as Transformers or Mamba. Furthermore, VDM aggregates local spatial features within each voxel’s 3D neighborhood, enabling the sequence models to more effectively utilize spatial positional cues during processing.

\subsection{Preliminaries}

\noindent \textbf{Transformer.} Transformers, based on self-attention mechanisms, have been widely adopted in both natural language processing and computer vision. The standard attention mechanism is defined as:
\begin{equation}
\text{Attention}(Q, K, V) = \text{softmax}\left( \frac{QK^\top}{\sqrt{d_k}} \right)V
\end{equation}
Here, $Q$ (Query), $K$ (Key), and $V$ (Value) represent the input projection vectors. While Transformers are powerful in capturing global dependencies, their quadratic time complexity poses a challenge when handling long sequences, such as in voxelized point clouds.

\noindent \textbf{State Space Models (SSMs).} SSMs model sequential data through continuous-time state equations:

\begin{equation}
h'(t) = Ah(t) + Bx(t)
\end{equation}
\begin{equation}
y(t) = Ch(t)
\end{equation}

where $x(t) \in \mathbb{R}^L$ is the continuous input signal, $y(t) \in \mathbb{R}^L$ is the output, and $A \in \mathbb{R}^{N \times N}$, $B \in \mathbb{R}^{N \times 1}$, and $C \in \mathbb{R}^{1 \times N}$ are learnable parameters. The discrete form is expressed as:

\begin{equation}
h_t = A'h_{t-1} + B'x_t
\end{equation}
\begin{equation}
y_t = Ch_t
\end{equation}

where $A'$ and $B'$ are obtained from continuous $A$ and $B$ via a discretization rule: $A' = f_A(\Delta, A)$ and $B' = f_B(\Delta, A, B)$. Mamba-based SSM models~\cite{gu2023mamba} have demonstrated significant advantages over Transformers in terms of modeling long-range dependencies and computational efficiency. In the context of 3D point cloud detection, such models~\cite{jin2025unimamba,liu2024lion} have shown promising results and even outperformed many Transformer-based architectures.

\noindent \textbf{Sparse Residual Block (SRB).} In CNN-based 3D detectors such as SAFDNet~\cite{SAFDnet}, the Sparse Residual Block (SRB) is commonly used for efficient feature extraction. In our design, VDM is composed of stacked SRB modules and additional SubM3D layers to enable both \textit{voxel expansion} and spatial feature aggregation. This structure facilitates the enrichment of voxel-level spatial representations before they are passed into the serialized backbone.

\subsection{Overall Architecture}

As illustrated in Figure~\ref{fig:framework}, the Voxel Densification Module (VDM) is designed to perform voxel-level densification of point clouds and aggregate fine-grained voxel features. This facilitates the generation of rich foreground voxel representations, serving as a beneficial foundation for subsequent Transformer-based and SSM-based detectors. 

Inspired by CNN-based paradigms, the proposed VDM is composed of alternating Submanifold 3D Convolution (SubM3D) layers and Sparse Residual Blocks (SRB). Additionally, two Sparse 3D Convolution (SPConv3D) layers are inserted to downsample the voxel feature maps, enabling hierarchical feature learning and effective \textit{spatial expansion}.
\begin{figure}[htbp]
    \centering
    \includegraphics[width=0.95\linewidth]{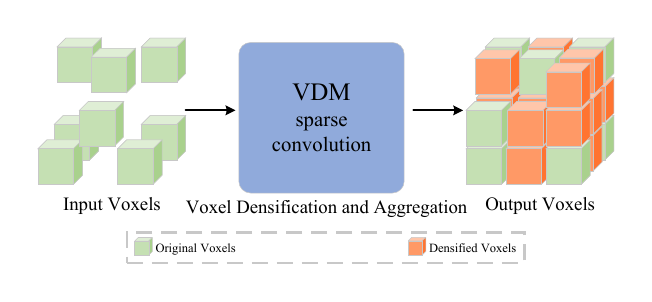} 
    \caption{Visualization of the Voxel Densification Module (VDM) applied to the original point clouds, demonstrating its effect in voxel-level expansion and feature aggregation.}
    \label{fig:diffusion}
\end{figure}

\noindent \textbf{Voxel Densification in VDM.} Voxel densification plays a significant role in enhancing model detection performance. Inspired by the property of convolution operations to increase the number of sparse features (dilation), we introduce a sparse 3D convolution with a kernel size of 3 within our proposed VDM. This module enables \textit{expansion} over the input voxel grid, thus enlarging the spatial distribution of informative voxels. Figure~\ref{fig:diffusion} illustrates how VDM expands the original point cloud into a denser voxel representation, providing a richer feature basis for subsequent serial detection networks. 
To quantitatively evaluate this effect, we perform a statistical analysis on the voxel count post-VDM processing, as shown in Table~\ref{tab:voxel_count_effect}. As observed, the number of total and foreground voxels significantly increases after applying VDM. This expansion ensures that subsequent modules—particularly those based on Transformers or SSMs—can access more comprehensive foreground point features, thereby improving object detection accuracy. It is worth noting that \textit{voxel expansion} serves as a crucial component of the VDM. Our later experiments will further highlight its contribution to performance gains.

\begin{figure}[tbp]
    \centering
    \includegraphics[width=0.95\linewidth]{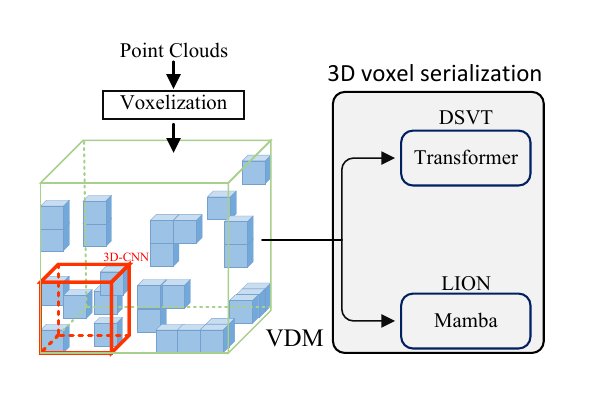} 
    \caption{Fine-grained feature aggregation within the VDM, utilizing convolutional operations to aggregate relative spatial information across voxels.}
    \label{fig:fine-grained}
\end{figure}

\noindent \textbf{Fine-Grained Feature Aggregation.} To capture intricate spatial details, we design VDM as a hierarchical framework. As shown in Figure~\ref{fig:framework}(a), the input sparse voxel grid is first processed by a Submanifold 3D Convolution (SubM-3D) stem to generate the initial feature embedding. Let $\mathbf{V}^{(0)} = (\mathcal{P}^{(0)}, \mathcal{F}^{(0)})$ denote this initial state. Here, $\mathcal{P}^{(0)} = \{p_i \mid p_i \in \mathbb{Z}^3\}_{i=1}^{N_0}$ represents the set of discrete coordinates for $N_0$ active voxels, and $\mathcal{F}^{(0)} \in \mathbb{R}^{N_0 \times C}$ denotes the corresponding feature vectors.

To achieve effective voxel densification and context aggregation, we subsequently employ a \textit{cascaded downsampling strategy} consisting of $L=2$ stages. Each stage $l$ comprises a sequence of Sparse Residual Blocks (SRB-3D) for feature refinement, followed by a strided Sparse 3D Convolution (SpConv-3D) for spatial expansion. Formally, we define the stage transformation $\mathbf{V}^{(l)} = \mathcal{H}_l(\mathbf{V}^{(l-1)})$ for $l \in \{1, 2\}$. The operator $\mathcal{H}_l$ is defined as the \textit{sequential composition} of two mapping functions:

First, the \textit{feature refinement} step utilizes the SRB-3D module, denoted as $\Psi(\cdot)$, to extract high-level semantics while maintaining the input topology:
\begin{equation}
\label{eq:refinement}
\hat{\mathcal{F}}^{(l-1)} = \Psi(\mathcal{F}^{(l-1)})
\end{equation}
where $\hat{\mathcal{F}}^{(l-1)}$ denotes the refined features at the current scale.

Second, the \textit{geometric expansion and aggregation} are performed simultaneously by the SpConv-3D layer with learnable weights $\mathcal{W}^{(l)}$ and a stride of $s=2$. This operation transforms both the geometry and features:
\begin{equation}
\label{eq:geometry}
\mathcal{P}^{(l)} = \left\{ \left\lfloor \frac{p + k}{2} \right\rfloor \;\middle|\; p \in \mathcal{P}^{(l-1)}, \, k \in \mathcal{K} \right\}
\end{equation}
\begin{equation}
\label{eq:aggregation}
\mathcal{F}^{(l)} = \mathcal{W}^{(l)} \ast \hat{\mathcal{F}}^{(l-1)}
\end{equation}
Here, $\mathcal{K}$ is the kernel support. Notably, despite the reduction in resolution due to the stride, the cardinality of the active voxel set $|\mathcal{P}^{(l)}|$ remains higher than that of a standard baseline at the equivalent voxel size. This phenomenon, attributed to the spatial diffusion effect of the kernel $\mathcal{K}$ in Equation~\ref{eq:geometry}, explicitly bridges the spatial-serial gap. The final output of VDM is the densified sparse tensor at the second stage, i.e., $\mathbf{V}' = \mathbf{V}^{(2)} = (\mathcal{P}^{(2)}, \mathcal{F}^{(2)})$, which provides the enriched representation (at $1/4$ scale) for serialization.

\noindent \textbf{Integration with SSMs and Transformer Modules.} The proposed VDM module can be integrated with both Mamba-based and Transformer-based models. By effectively diffusing sparse voxels and aggregating fine-grained point cloud voxel features, VDM further enhances the detection accuracy of the overall model. For the 3D feature maps processed by VDM, we first serialize them into one-dimensional sequences along the $x$ and $y$ directions. As the sequence length increases, Mamba-based networks are able to adopt larger grouping sizes due to their lower computational complexity compared to Transformer-based counterparts. Given the strong performance of LION~\cite{liu2024lion} and DSVT~\cite{wang2023dsvt} in the field of 3D object detection, we integrate the VDM module into both networks and conduct experiments to validate its effectiveness.

\begin{table}[tbp]
\centering

\resizebox{\columnwidth}{!}{
\begin{tabular}{c|c|c|c}
\hline
\textbf{Method} & \textbf{Overall Voxel Count} & \textbf{Foreground Voxel Count} & \textbf{mAPH L2} \\
\hline
Non(LION) & 36340 & 2616 & 74.0 \\
VDM  & 188562 & 9146 & 74.8 \\
VDM-OD  & 216601 & 10705 & 74.8 \\
\hline
\end{tabular}
}
\caption{Effect of VDM on voxel counts in input point clouds on the Waymo validation dataset. The overall and foreground voxel counts are computed as the total number of corresponding voxels across all frames divided by the total number of frames.}
\label{tab:voxel_count_effect}
\end{table}

\begin{table*}[ht]
\centering

\setlength{\tabcolsep}{1.1mm}
\begin{tabular}{c|c|>{\centering\arraybackslash}p{1.4cm}|>{\centering\arraybackslash}p{1.4cm}|>{\centering\arraybackslash}p{1.4cm}|>{\centering\arraybackslash}p{1.4cm}|>{\centering\arraybackslash}p{1.4cm}|>{\centering\arraybackslash}p{1.4cm}|c}
\hline
\multirow{2}{*}{Methods} & \multirow{2}{*}{Operator} & \multicolumn{2}{c|}{Vehicle 3D AP/APH} & \multicolumn{2}{c|}{Pedestrian 3D AP/APH} & \multicolumn{2}{c|}{Cyclist 3D AP/APH} & \multirow{2}{*}{\parbox{2cm}{\centering mAP/mAPH\\(L2)}} \\
&& L1 & L2 & L1 & L2 & L1 & L2 & \\
\hline
SECOND \cite{yan2018second} & \multirow{11}{*}{\centering\rotatebox[origin=c]{90}{CNN}} & 72.3/71.7 & 63.9/63.3 & 68.7/58.2 & 60.7/51.3 & 60.6/59.3 & 58.3/57.0 & 61.0/57.2 \\
PointPillars \cite{lang2019pointpillars} &  & 72.1/71.5 & 63.6/63.1 & 70.6/56.7 & 62.8/50.3 & 64.4/62.3 & 61.9/59.9 & 62.8/57.8 \\
CenterPoint \cite{centerpoint}&&74.2/73.6&66.2/65.7&76.6/70.5&68.8/63.2&72.3/71.1&69.7/68.5&68.2/65.8 \\
PillarNet-18 \cite{pillarnet}&&78.2/77.7&70.4/69.9&79.8/72.6&71.6/64.9&70.4/69.3&67.8/66.7&69.9/67.2 \\
FSD \cite{fan2022fullyfsd}&&79.2/78.8&70.5/70.1&82.6/77.3&73.9/69.1&77.1/76.0&74.4/73.3&72.9/70.8 \\
AFDetV2 \cite{hu2022afdetv2}&&77.6/77.1&69.7/69.2&80.2/74.6&72.2/67.0&73.7/72.7&71.0/70.1&71.0/68.8 \\
VoxelNext \cite{voxelnext}&&78.2/77.7&69.9/69.4&81.5/76.3&73.5/68.6&76.1/74.9&73.3/72.2&72.2/70.1 \\
CenterFormer \cite{zhou2022centerformer}&&75.0/74.4&69.9/69.4&78.6/73.0&73.6/68.3&72.3/71.3&69.8/68.8&71.1/68.9 \\
PV-RCNN++ \cite{shi2023pvrcnn++}&&79.3/78.8&70.6/70.2&81.3/76.3&73.2/68.0&73.7/72.7&71.2/70.2&71.7/69.5 \\
HEDNet \cite{hednet}&&81.1/80.6&73.2/72.7&84.4/80.0&76.8/72.6&78.7/77.7&75.8/74.9&75.3/73.4 \\
VPF \cite{huang2024voxelvpf}&&80.2/79.7&71.9/71.5&82.5/76.9&74.8/69.4&77.1/76.0&74.2/73.2&73.6/71.4 \\
\hline
TransFusion \cite{bai2022transfusion} & \multirow{6}{*}{\centering\rotatebox[origin=c]{90}{Transformer}} & –/– & –/65.1 & –/– & –/63.7 & –/– & –/65.9 & –/64.9 \\
FlatFormer \cite{liu2023flatformer} && - & 69.0/68.6 & - & 71.5/65.3 & –/– & 68.6/67.5 & 69.7/67.2 \\
OcTr \cite{zhou2023octr} && 78.1/77.6 & 69.8/69.3 & 80.8/74.4 & 72.5/66.5 & 72.6/71.5 & 69.9/68.9 & 70.7/68.2 \\
DSVT-Pillar \cite{wang2023dsvt} && 79.3/78.8 & 70.9/70.5 & 82.8/77.0 & 75.2/69.8 & 76.4/75.4 & 73.6/72.7 & 73.2/71.0 \\
DSVT-Voxel \cite{wang2023dsvt} && 79.2/78.8 & 70.9/70.5 & 82.9/77.7 & 75.2/70.3 & 77.2/76.2 & 74.3/73.4 & 73.5/71.4 \\
VDM-DSVT(Ours)&&\textbf{79.7/79.3}&\textbf{71.4/71.0}&\textbf{84.3/79.7}&\textbf{76.7/72.3}&\textbf{78.4/77.4}&\textbf{75.6/74.6}&\textbf{74.6/72.6}\\
\hline
Voxel Mamba \cite{voxel-mamba} & \multirow{5}{*}{\centering\rotatebox[origin=c]{90}{SSMs}} & 80.8/80.3 & 72.6/72.2 & 85.0/80.8 & 77.7/73.6 & 78.6/77.6 & 75.7/74.8 & 75.3/73.6 \\
LION \cite{liu2024lion} && 80.3/79.9 & 72.0/71.6 & 85.8/81.4 & 78.5/74.3 & 80.1/79.0 & 77.2/76.2 & 75.9/74.0 \\
UniMamba \cite{jin2025unimamba} && 80.6/80.1& 72.3/71.8& 86.0/81.3 &78.7/74.1& 80.3/79.3 &77.5/76.5& 76.1/74.1\\
VDM-Mamba (Ours)&&81.0/80.5&72.6/72.2&\textbf{86.4/82.5}&\textbf{79.3/75.5}&80.4/79.4&77.5/76.5&76.5/\textbf{74.8}\\
VDM-OD (Ours)&&\textbf{81.3/80.9}&\textbf{73.0/72.6}&86.1/82.1&78.9/75.0&\textbf{80.7/79.7}&\textbf{77.8/76.9}&\textbf{76.6/74.8}\\
\hline
\end{tabular}
\caption{
Comparison with prior methods (CNNs, Transformers, and SSMs) on the Waymo Open Dataset \textbf{validation set} using 100\% training data. Metrics are reported as mAP/mAPH (\%) $\uparrow$ for overall results and AP/APH (\%) $\uparrow$ per category. All models are trained under the single-frame setting. “–” indicates unavailable results. VDM-OD denotes a variant where the \textbf{Only-Densification} structure is used within the VDM framework. The best-performing results are highlighted in bold.
}

\label{tab:waymo}
\end{table*}

\begin{table*}[htbp]
\centering

\begin{tabular}{c|cc|cccccccccc}
\hline
\noalign{\vskip 0.5mm}
\multicolumn{13}{c}{\textit{Performances on the validation set}}\\
\hline
\noalign{\vskip 0.5mm}
Method & NDS & mAP & Car & Truck & Bus & T.L. & C.V. & Ped. & M.T. & Bike & T.C. & B.R. \\
\hline
\noalign{\vskip 0.5mm}
CenterPoint \cite{centerpoint} & 66.5 & 59.2 & 84.9 & 57.4 & 70.7 & 38.1 & 16.9 & 85.1 & 59.0 & 42.0 & 69.8 & 68.3 \\
Transfusion-LiDAR \cite{bai2022transfusion} & 70.1 & 65.5 & 86.9 & 60.8 & 73.1 & 43.4 & 25.2 & 87.5 & 72.9 & 57.3 & 77.2 & 70.3 \\
DSVT \cite{wang2023dsvt} & 71.1 & 66.4 & 87.4 & 62.6 & 75.9 & 42.1 & 25.3 & 88.2 & 74.8 & 58.7 & 77.9 & 71.0 \\
HEDNet \cite{hednet} & 71.4 & 66.7 & 87.7 & 60.6 & 77.8 & 50.7 & 28.9 & 87.1 & 74.3 & 56.8 & 76.3 & 66.9 \\
SAFDNet \cite{SAFDnet} & 71.0 & 66.3 & 87.6 & 60.8 & 78.0 & 43.5 & 26.6 & 87.8 & 75.5 & 58.0 & 75.0 & 69.7 \\
LION \cite{liu2024lion} & 72.1 & 68.0 & 87.9 & 64.9 & 77.6 & 44.4 & 28.5 & 89.6 & 75.6 & 59.4 & 80.8 & 71.6 \\
FSHNet\cite{liu2025fshnet}& 71.7 &68.1& 88.7& 61.4& 79.3& 47.8& 26.3& 89.3& \textbf{76.7}& 60.5& 78.6& 72.3
\\
UniMamba \cite{jin2025unimamba}   &72.6&68.5 &88.7& 64.7& \textbf{79.7}& \textbf{47.9}& 28.7& 89.7& 74.6& 59.1& 79.5& 72.3\\

\hline
\noalign{\vskip 0.5mm}
VDM-mamba(Ours) &72.5&68.1&88.5&59.5&78.8&43.2&\textbf{30.6}&\textbf{90.2}&75.2&\textbf{61.1}&\textbf{81.4}&\textbf{72.5}\\
VDM-OD (Ours) &\textbf{72.9}&\textbf{68.5}&\textbf{88.7}&\textbf{65.7}&78.9&43.8&29.7&90.2&75.7&60.9&81.1&70.3\\
\noalign{\vskip 0.5mm} \hline
\multicolumn{13}{c}{\textit{Performances on the test set}}\\
\hline
\noalign{\vskip 0.5mm}
TransFusion-LiDAR \cite{bai2022transfusion} & 70.2 & 65.5 & 86.2 & 56.7 & 66.3 & 58.8 & 28.2 & 86.1 & 68.3 & 44.2 & 82.0 & 78.2 \\
DSVT \cite{wang2023dsvt} & 72.7 & 68.4 & 86.8 & 58.4 & 67.3 & 63.1 & 37.1 & 88.0 & 73.0 & 47.2 & 84.9 & 78.4 \\
HEDNet \cite{hednet} & 72.0 & 67.7 & 87.1 & 56.5 & 70.4 & 63.5 & 33.6 & 87.9 & 70.4 & 44.8 & 85.1 & 78.1 \\
VPF \cite{huang2024voxelvpf} & 72.7 & 67.0 & 85.8 & 55.1 & 63.5 & 62.1 & 33.3 & 87.6 & 72.5 & 48.6 & 82.9 & 78.2 \\
SAFDNet \cite{SAFDnet} & 72.3 & 68.3 & 87.3 & 57.3 & 68.0 & 63.7 & \textbf{37.3} & 89.0 & 71.1 & 44.8 & 84.9 & 79.5 \\
LION \cite{liu2024lion}&73.9& 69.8& 87.2& 61.1& 68.9& 65.0& 36.3& 90.0& 74.0& 49.2& 87.3& 79.5\\
UniMamba\cite{jin2025unimamba} & \textbf{74.0} & 70.2 & 87.9 & 60.4 & \textbf{70.9} & 65.9 & 36.7 & 90.5 & 73.5 & 49.5 & 86.9 & 79.4 \\
\hline
\noalign{\vskip 0.5mm}
VDM-mamba(Ours)&73.7&70.0&88.0&60.1&67.6&\textbf{66.0}&35.1&\textbf{90.9}&72.2&50.9&\textbf{89.2}&\textbf{80.0}\\

VDM-OD (Ours) & 73.9 & \textbf{70.5} & \textbf{88.3} & \textbf{62.2} & 70.7 & 65.3 & 35.0 & 90.6 & \textbf{74.4} & \textbf{51.1} & 88.3 & 78.8 \\
\noalign{\vskip 0.5mm}

\hline
\end{tabular}
\caption{Performances on the nuScenes validation and test set. ‘T.L.’, ‘C.V.’, ‘Ped.’, ‘M.T.’, ‘T.C.’, and ’B.R.’ are short for trailer, construction vehicle, pedestrian, motor, traffic cone, and barrier, respectively. VDM-OD denotes a variant where the Only-Densification
structure is used within the VDM framework. }
\label{nuscenes}
\end{table*}

\begin{table*}[ht]
\centering
\setlength{\tabcolsep}{0.8mm}  
{\scriptsize
\begin{tabular}{c|c|rrrrrrrrrrrrrrrrrrrrrrrrrrr}
\hline
\noalign{\vskip 0.5mm}
Method & \rotatebox{90}{mAP} & \rotatebox{90}{Vehicle} & \rotatebox{90}{Bus} & \rotatebox{90}{Pedestrian} & \rotatebox{90}{Stop Sign} & \rotatebox{90}{Box Truck} & \rotatebox{90}{Bollard} & \rotatebox{90}{C-Barrel} & \rotatebox{90}{Motorcyclist} & \rotatebox{90}{MPC-Sign} & \rotatebox{90}{Motorcycle} & \rotatebox{90}{Bicycle} & \rotatebox{90}{A-Bus} & \rotatebox{90}{School Bus} & \rotatebox{90}{Truck Cab} & \rotatebox{90}{C-Cone} & \rotatebox{90}{V-Trailer} & \rotatebox{90}{Sign} & \rotatebox{90}{Large Vehicle} & \rotatebox{90}{Stroller} & \rotatebox{90}{Bicyclist} & \rotatebox{90}{Truck} & \rotatebox{90}{MBT} & \rotatebox{90}{Dog} & \rotatebox{90}{Wheelchair} & \rotatebox{90}{W-Device} & \rotatebox{90}{W-Rider} \\
\hline
CenterPoint \cite{centerpoint} & 22.0 & 67.6 & 38.9 & 46.5 & 16.9 & 37.4 & 40.1 & 32.2 & 28.6 & 27.4 & 33.4 & 24.5 & 8.7 & 25.8 & 22.6 & 29.5 & 22.4 & 6.3 & 3.9 & 0.5 & 20.1 & 22.1 & 0.0 & 3.9 & 0.5 & 10.9 & 4.2 \\
HEDNet \cite{hednet} & 37.1 & 78.2 & 47.7 & 67.6 & 46.4 & 45.9 & 56.9 & 67.0 & 48.7 & 46.5 & 58.2 & 47.5 & 23.3 & 40.9 & 27.5 & 46.8 & 27.9 & 20.6 & 6.9 & 27.2 & 38.7 & 21.6 & 0.0 & 30.7 & 9.5 & 28.5 & 8.7 \\
VoxelNeXt \cite{voxelnext} & 30.7 & 72.7 & 38.8 & 63.2 & 40.2 & 40.1 & 53.9 & 64.9 & 44.7 & 39.4 & 42.4 & 40.6 & 20.1 & 25.2 & 19.9 & 44.9 & 20.9 & 14.9 & 6.8 & 15.7 & 32.4 & 16.9 & 0.0 & 14.4 & 0.1 & 17.4 & 6.6 \\
FSDV2 \cite{fan2024fsdv2} & 37.6 & 77.0 & 47.6 & 70.5 & 43.6 & 41.5 & 53.9 & 58.5 & 56.8 & 39.0 & 60.7 & 49.4 & 28.4 & 41.9 & 30.2 & 44.9 & 33.4 & 16.6 & 7.3 & 32.5 & 45.9 & 24.0 & 1.0 & 12.6 & 17.1 & 26.3 & 17.2 \\
SAFDNet \cite{SAFDnet} & 39.7 & 78.5 & 49.4 & 70.7 & 51.5 & 44.7 & 65.7 & 72.3 & 54.3 & 49.7 & 60.8 & 50.0 & 31.3 & 44.9 & 24.7 & 55.4 & 31.4 & 22.1 & 7.1 & 31.1 & 42.7 & 23.6 & 0.0 & 26.1 & 1.4 & 30.2 & 11.5 \\
LION \cite{liu2024lion} & 41.5 & 75.1 & 43.6 & 73.9 & 53.9 & 45.1 & 66.4 & 74.7 & 61.3 & 48.7 & 65.1 & 56.2 & 21.7 & 42.7 & 25.3 & 58.4 & 28.9 & 23.6 & 8.3 & 49.5 & 47.3 & 19.0 & 0.0 & 31.4 & 8.7 & 37.6 & 11.8 \\
FSHNet\cite{liu2025fshnet}& 40.2& 77.4& 48.3& 72.9& 50.0& 47.2& 63.4& 69.9& 56.1& 43.8& 62.5& 53.7& 31.8& 44.8& 28.9& 56.6& 32.0& 23.0& 8.7& 34.5& 41.0\\
UniMamba \cite{jin2025unimamba} & 42.0 & 78.9 & 47.9 & 74.3 & 51.8 & 46.8 & 67.8 & 76.9 & 55.8 & 51.7 & 62.8 & 52.2 & 30.2 & 44.6 & 24.6 & 28.1 & 59.4 & 32.2 & 23.2 & 6.7 & 41.5 & 48.5 & 0.0 & 26.4 & 8.1 & 36.4 & 13.7 \\
\hline
VDM (Ours) & 42.3 &78.2&48.2&75.2&53.2&45.7&68.2&77.4&60.3&47.2&67.0&58.0&24.9&43.0&24.9&60.1&31.4&25.1&7.8&42.5&46.3&20.9&0.0&29.7&10.3&39.8&13.6\\
VDM-OD (Ours)&\textbf{42.6}&79.0&49.5&75.9&54.5&45.5&69.9&77.9&56.9&47.6&66.6&59.2&24.0&44.2&24.7&60.9&33.2&24.9&7.5&39.5&51.3&20.1&0.0&33.7&7.8&40.1&12.5\\
\hline
\end{tabular}
\caption{
Comparison with prior methods on the Argoverse2 validation set. ‘Vehicle’, ‘C-Barrel’, ‘MPC-Sign’, ‘A-Bus’, ‘C-Cone’, ‘V-Trailer’, ‘MBT’, ‘W-Device’, and ‘W-Rider’ are abbreviations for regular vehicle, construction barrel, mobile pedestrian crossing sign, articulated bus, construction cone, vehicular trailer, message board trailer, wheeled device, and wheeled rider, respectively. \textbf{VDM} and \textbf{VDM-OD} refer to the proposed \textit{VDM-Mamba} and \textit{VDM-OD} variants.
}
\label{argov2}
}
\end{table*}

\section{Experiments}
\subsection{Datasets and Evaluation Metrics}

\noindent \textbf{Waymo Open Dataset.} The Waymo Open Dataset (WOD)~\cite{sun2020scalabilitywaymo} includes 1,150 scenes, divided into 798 for training, 202 for validation, and 150 for testing. Each scene covers a perception range of $150\,\text{m} \times 150\,\text{m}$. For evaluation, WOD adopts 3D mean Average Precision (mAP) and mAP weighted by heading accuracy (mAPH). Each metric includes two difficulty levels: L1 for objects with more than five points, and L2 for bounding boxes containing one to five points.

\noindent \textbf{nuScenes Dataset.} The nuScenes dataset~\cite{caesar2020nuscenes} contains 1,000 scenes, with 750 for training, 150 for validation, and 150 for testing. It uses mean Average Precision (mAP) and nuScenes Detection Score (NDS) as evaluation metrics. The dataset covers 10 annotated object classes, representing most common categories in traffic scenarios.

\noindent \textbf{Argoverse 2 Dataset.} The Argoverse 2 dataset~\cite{chang2019argoverse} comprises 1,000 sequences, with 700 for training, 150 for validation, and 150 for testing. The perception range extends up to 200 meters. It uses mean Average Precision (mAP) as the evaluation metric.

\noindent \textbf{ONCE Dataset.} The ONCE dataset~\cite{mao2021once} consists of 5,000, 3,000, and 8,000 frames for training, validation, and testing respectively. It defines three detection classes: Vehicle, Pedestrian, and Cyclist. The final performance is evaluated using mean Average Precision (mAP) across all three classes.

\subsection{Implementation Details}

\noindent \textbf{Network Architecture.}  
We evaluate the proposed VDM module by integrating it with both the SSMs-based LION~\cite{liu2024lion} and the Transformer-based DSVT~\cite{wang2023dsvt} models. We use voxel sizes of $(0.075\,\text{m}, 0.075\,\text{m}, 0.25\,\text{m})$, $(0.08\,\text{m}, 0.08\,\text{m}, 0.1875\,\text{m})$, $(0.1\,\text{m}, 0.1\,\text{m}, 0.25\,\text{m})$, and $(0.1\,\text{m}, 0.1\,\text{m}, 0.25\,\text{m})$ for the nuScenes, Waymo, Argoverse 2, and ONCE datasets respectively. The number of voxel feature channels is set to 64. Specifically for the nuScenes dataset, we incorporate a Sparse Residual Module to enhance spatial feature interactions among voxels.


In the VDM module, the SubM3D (including SRB) layers use a kernel size of 3. All SparseConv3D layers adopt a kernel size of 3, stride of 2, and padding of 1 to downsample the feature map resolution. The channel configurations within VDM are set as $(64, 32)$, $(32, 64)$, and $(64, 128)$. For LION and DSVT, we follow the default parameter settings from their original papers without modification. Additionally, we conduct an \textit{only-densification} experiment, in which the voxel size along the $x$ and $y$ axes is scaled by a factor of 4, while the $z$ axis remains unchanged.

\noindent \textbf{Training Process.}  
We train all models using 8 NVIDIA A40 GPUs. On the Waymo Open Dataset, the VDM-SSMs configuration is trained for 24 epochs with a batch size of 16, while the VDM-DSVT configuration uses 24 epochs with a batch size of 8. For the remaining three datasets, we adopt the same training strategy as that used in the LION~\cite{liu2024lion} baseline.

\subsection{Main Results}

To verify the effectiveness of the proposed module, we design two models: VDM-SSMs based on LION and VDM-Transformer based on DSVT. We evaluate their performance on four benchmark datasets: Waymo, nuScenes, Argoverse V2, and ONCE.

\noindent \textbf{Results on WOD.} As shown in Table~\ref{tab:waymo}, we incorporate the VDM module into both Mamba-based and Transformer-based models. Among them, VDM-Mamba achieves a state-of-the-art (SOTA) result of 74.8 mAPH (L2), outperforming the corresponding baseline LION model (without VDM) by +0.8 mAPH, and also surpassing the recent UniMamba~\cite{jin2025unimamba}. In addition, VDM-DSVT also demonstrates performance gains, achieving a +1.2 mAPH (L2) improvement over the original DSVT, which validates the general applicability of the VDM module to both SSM-based and Transformer-based detectors.

When only the densification mechanism is retained in VDM (i.e., VDM-OD), the model still achieves the best results on the Vehicle and Cyclist categories, highlighting the importance of the densification component in the overall design. Moreover, VDM-Mamba achieves the best performance in the Pedestrian category, suggesting that the fine-grained aggregation ability introduced by VDM is particularly beneficial for small object detection.

\begin{table*}[ht]
\centering
\setlength{\tabcolsep}{1.6mm}  
{\small
\begin{tabular}{c|c|c|c|c|c|c|c|c|c|c|c|c|c}
\hline
\noalign{\vskip 0.5mm}
\multirow{2}{*}{Method} & \multicolumn{4}{c|}{Vehicle}  & \multicolumn{4}{c|}{Pedestrian}  & \multicolumn{4}{c|}{Cyclist} & \multirow{2}{*}{mAP} \\
& overall & 0-30m & 30-50m & 50m+ & overall & 0-30m & 30-50m & 50m+ & overall & 0-30m & 30-50m & 50m+ & \\
\noalign{\vskip 0.5mm}
\hline
PointRCNN \cite{shi2019pointrcnn} & 52.1 & 74.5 & 40.9 & 16.8 & 4.3 & 6.2 & 2.4 & 0.9 & 29.8 & 46.0 & 20.9 & 5.5 & 28.7 \\
PointPillars \cite{lang2019pointpillars} & 68.6 & 80.9 & 62.1 & 47.0 & 17.6 & 19.7 & 15.2 & 10.2 & 46.8 & 58.3 & 40.3 & 25.9 & 44.3 \\
SECOND \cite{yan2018second} & 71.2 & 84.0 & 63.0 & 47.3 & 26.4 & 29.3 & 24.1 & 18.1 & 58.0 & 70.0 & 52.4 & 34.6 & 51.9 \\
PV-RCNN \cite{shi2020pvrcnn} & 77.8 & 89.4 & 72.6 & 58.6 & 23.5 & 25.6 & 22.8 & 17.3 & 59.4 & 71.7 & 52.6 & 36.2 & 53.6 \\
CenterPoint \cite{centerpoint} & 66.8 & 80.1 & 59.6 & 43.4 & 49.9 & 56.2 & 42.6 & 26.3 & 63.5 & 74.3 & 57.9 & 41.5 & 60.1 \\
PointPainting \cite{vora2020pointpainting} & 66.2 & 80.3 & 59.8 & 42.3 & 44.8 & 52.6 & 36.6 & 22.5 & 62.3 & 73.6 & 57.2 & 40.4 & 57.8 \\
LION \cite{liu2024lion} & 78.2 & 89.1 & 72.6 & 57.5 & 53.2 & 62.4 & 44.0 & 24.5 & 68.5 & 79.2 & 63.2 & 47.1 & 66.6 \\
\hline
VDM-mamba (Ours) &78.6&89.1&74.3&59.3&54.2&63.3&43.2&26.1&69.9&79.4&64.9&49.8&\textbf{67.6}\\
VDM-OD (Ours) & 78.5 & 88.8 & 73.1 & 59.7 & 51.0 & 59.2 & 42.3 & 24.5 & 68.7 & 79.1 & 64.1 & 48.9 & 66.1 \\
\noalign{\vskip 0.5mm}
\hline
\end{tabular}
\caption{Comparison with previous methods on ONCE validation set.}
\label{once}
}
\end{table*}

\noindent \textbf{Results on nuScenes.} To further verify the effectiveness of the proposed VDM, we conduct experiments on both the nuScenes validation and test sets. As shown in Table~\ref{nuscenes}, on the validation set, VDM-Mamba achieves 68.1 mAP and 72.5 NDS, while the \textbf{VDM-OD} variant achieves even higher performance with \textbf{68.5 mAP} and \textbf{72.9 NDS}. Most notably, on the \textbf{test set}, VDM-OD achieves \textbf{70.5 mAP}, surpassing the previous state-of-the-art UniMamba (70.2 mAP) and LION (69.8 mAP), while maintaining a competitive 73.9 NDS.

Detailed analysis reveals that VDM-Mamba excels in detecting vulnerable road users, achieving the best results in the \textit{Pedestrian} and \textit{Bicycle} categories on the validation set. This demonstrates that the fine-grained aggregation component in the full VDM offers clear advantages for small object localization. However, the superior overall performance of VDM-OD (70.5 mAP on Test) suggests that the \textit{densification-only} structure plays the most critical role in mitigating feature sparsity and boosting general detection accuracy.

\noindent \textbf{Results on Argoverse V2.} As shown in Table~\ref{argov2}, our proposed models, VDM-Mamba and VDM-OD, achieve state-of-the-art detection performance on the Argoverse 2 validation set, reaching 42.3 mAP and 42.6 mAP, respectively. Compared with the strong baseline LION~\cite{liu2024lion}, VDM-Mamba yields a notable improvement of 0.8 mAP. Notably, VDM-OD, which retains only the densification structure within the VDM module, still surpasses all previous approaches, highlighting the effectiveness of the densification mechanism. These results further demonstrate the strong representational capacity and scalability of the VDM module in large-scale point cloud scenarios.

\begin{table*}[htbp]
\centering
\begin{tabular}{c|c|c|c|c|c|c|c|c}
\hline
\multirow{2}{*}{Fine-grained} & \multicolumn{4}{c|}{ONCE (Overall)}&\multicolumn{4}{c}{Waymo (Cyclist)} \\ 
  &  Vehicle & Pedestrian  &Cyclist &mAP  &AP (L1)&APH (L1)&AP (L2) & APH (L2)\\ \hline
Base (VDM-OD)  & 78.5 & 51.0&68.7&66.1 & 78.5&77.6&75.7&74.7\\ 
Fine (VDM)  & 78.6 & 54.2&69.9&67.6 &79.1&78.1&76.2&75.2\\ \hline
\end{tabular}
\caption{Ablation study on the ONCE and Waymo (20\% data) dataset for fine-grained feature aggregation. \textit{Base} denotes the detection results obtained by combining the VDM and Mamba network without fine voxel partitioning. \textit{Fine} represents the results after applying fine-grained voxel partitioning.}

\label{frain-comparsion}
\end{table*}

\begin{table}[ht]
\centering
\begin{tabular}{c|c|c|c|c}
\hline
\multirow{2}{*}{Method} & \multicolumn{4}{c}{Speed (s)}  \\ 
 &Waymo & nuScenes &Argoverse2  &ONCE  \\ \hline
LION & 0.206 &0.213  & 0.185 &0.149 \\ 
VDM & 0.357 & 0.291 & 0.320 & 0.228\\ \hline
\end{tabular}
\caption{Runtime comparison on different datasets. The inference speed is measured in seconds (s) on a single NVIDIA 3090 GPU with a batch size of 1.}
\label{runtime}
\end{table}

\begin{table}[ht]
\centering
\begin{tabular}{c|c|c|c}
\hline
\multirow{2}{*}{Densification} & \multicolumn{3}{c}{Waymo(3D AP/APH)L2}  \\ 
 &Vehicle & Pedestrian &Cyclist    \\ \hline
No(LION) & 68.8/68.3 & 77.3/72.6 & 74.7/73.8  \\ 
Yes(VDM-OD) & 70.3/69.9 & 78.2/73.8 & 76.0/75.0 \\ \hline
\end{tabular}
\caption{Detection results of the voxel densification module on the Waymo validation set (20\% data). ``No'' indicates that the original voxels are not diffused, and the detection results are entirely determined by LION \cite{liu2024lion}. ``Yes'' indicates that we only adopt the densification module from VDM, i.e., using sparse 3D spconv with a kernel size of 3 and a stride of 1.}
\label{ablity-diffusion}
\end{table}

\noindent \textbf{Results on ONCE.} 
VDM-Mamba demonstrates superior performance on the ONCE dataset (Table~\ref{once}). It achieves detection accuracies of 78.6 mAP, 54.2 mAP, and 69.9 mAP on the \textit{Vehicle}, \textit{Pedestrian}, and \textit{Cyclist} categories, respectively. The overall detection accuracy reaches \textbf{67.6 mAP}, representing a +1.0 mAP improvement over the baseline LION detector, establishing VDM-Mamba as the new state-of-the-art (SOTA).

In contrast to the nuScenes results, the \textbf{VDM-OD} variant (utilizing only densification) achieves \textbf{66.1 mAP}, which is lower than the full VDM framework. This performance gap highlights the critical contribution of the fine-grained feature aggregation component on this dataset. While voxel expansion addresses sparsity, the aggregation module captures detailed local geometric cues that prove essential for the ONCE benchmarks, validating the necessity of our dual-path design (densification + aggregation) for robust detection across diverse scenarios.

\begin{figure*}[ht]
    \centering
    \subfloat{\includegraphics[width=0.49\textwidth, height=5cm]{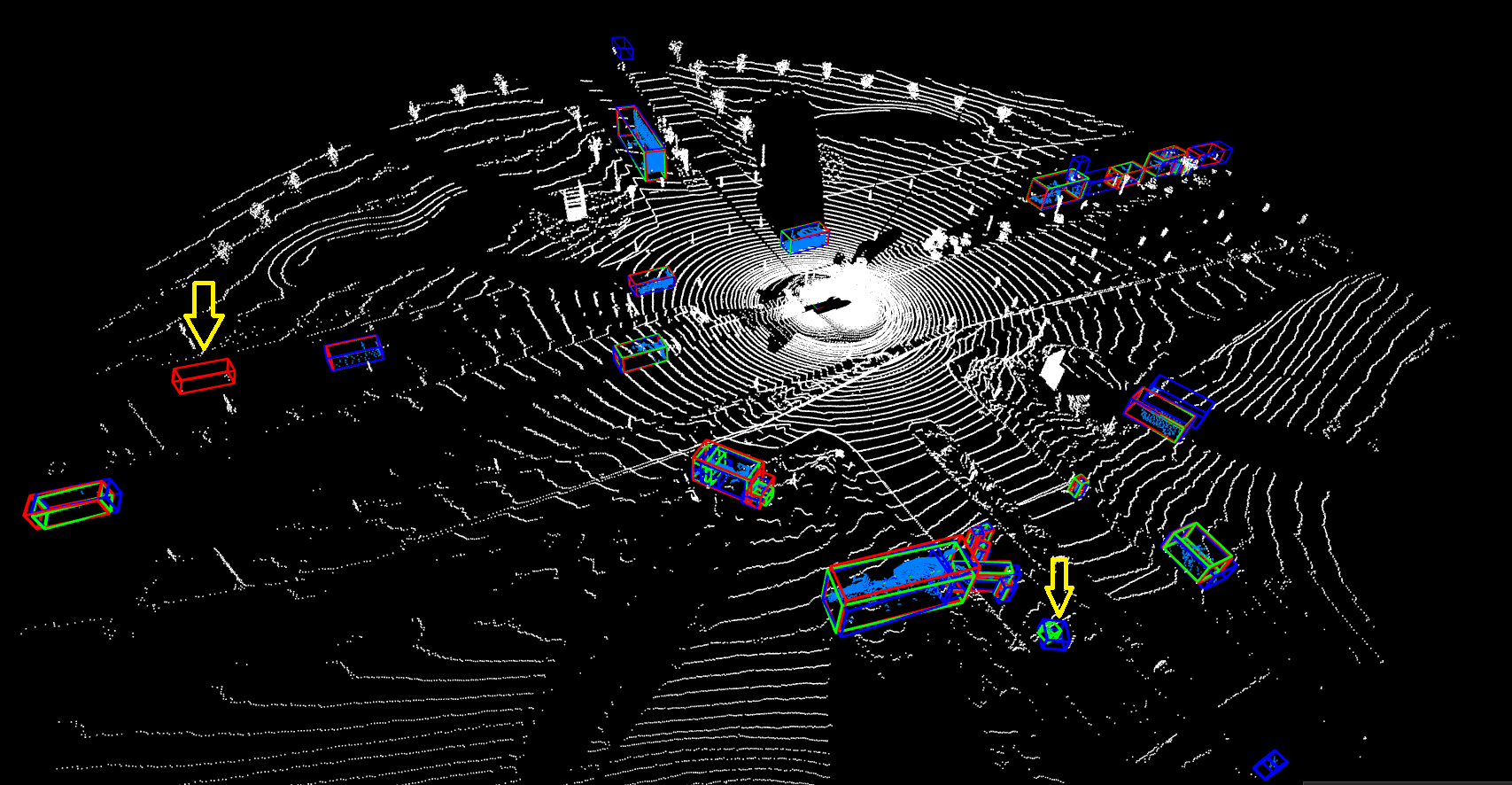}\label{fig:sub1}}
    \hfill
    \subfloat{\includegraphics[width=0.49\textwidth, height=5cm]{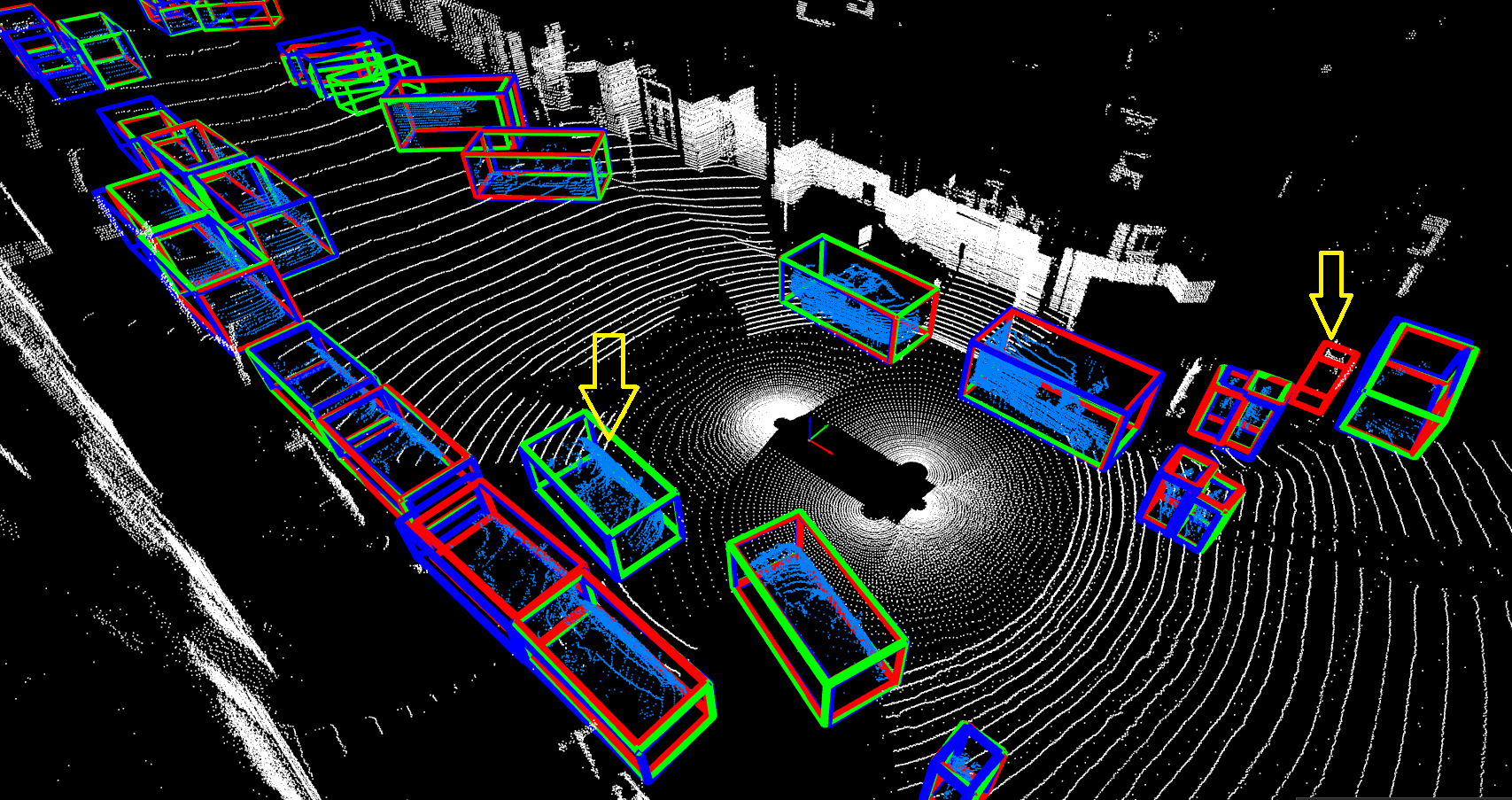}\label{fig:sub2}}
    
    
    \subfloat{\includegraphics[width=0.49\textwidth, height=5cm]{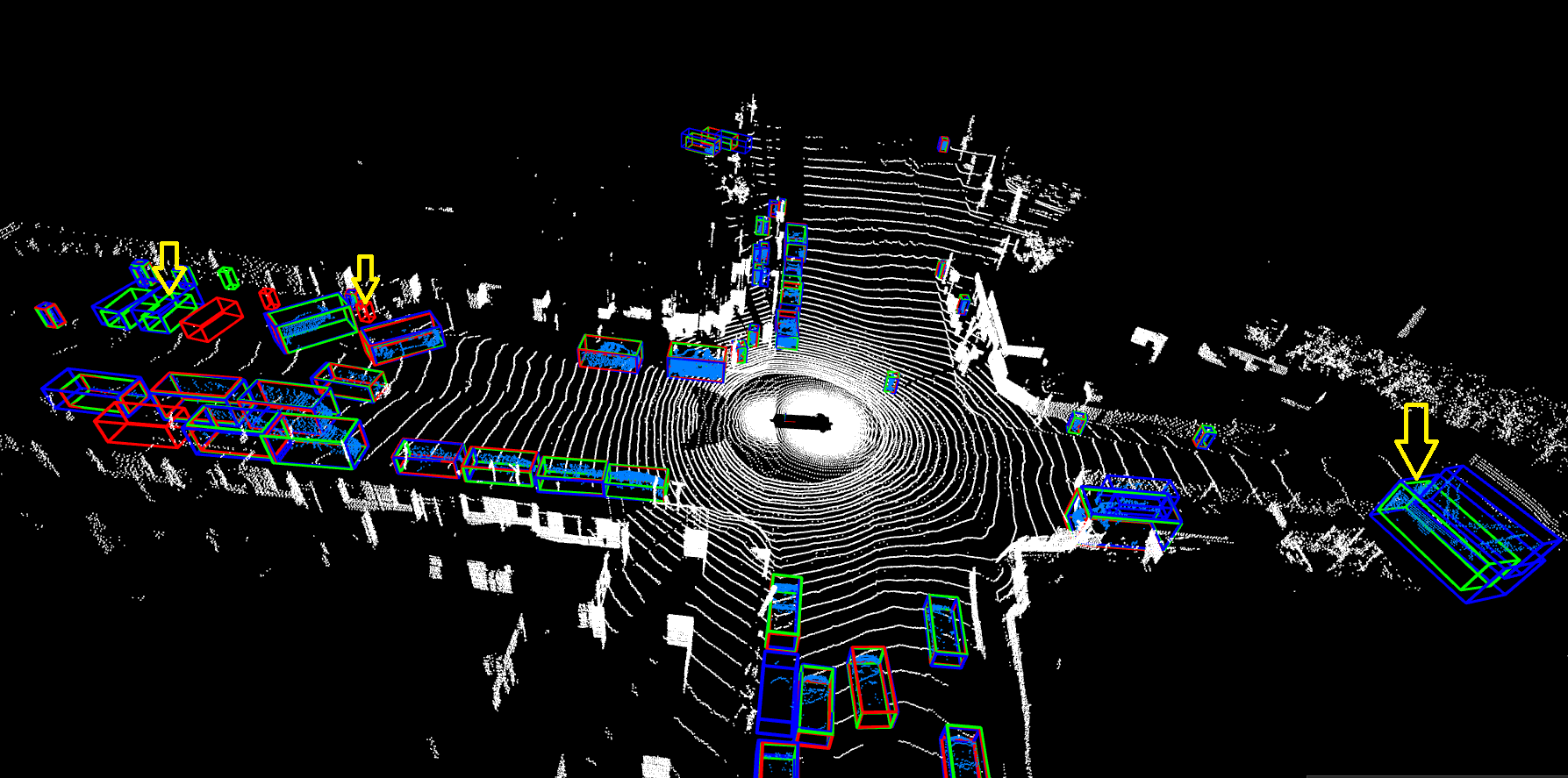}\label{fig:sub3}}
    \hfill
    \subfloat{\includegraphics[width=0.49\textwidth, height=5cm]{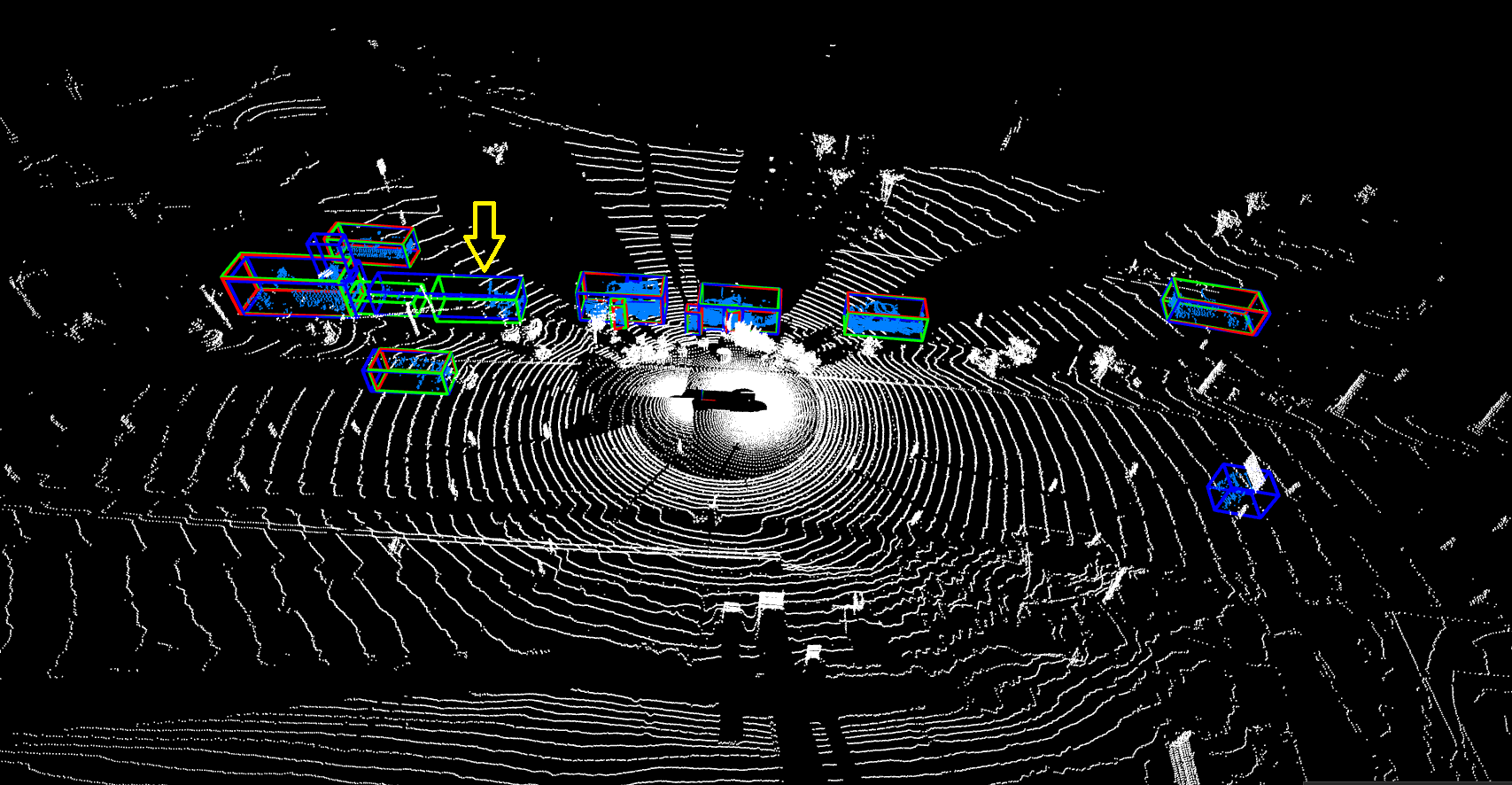}\label{fig:sub4}}

    \subfloat{\includegraphics[width=0.49\textwidth, height=5cm]{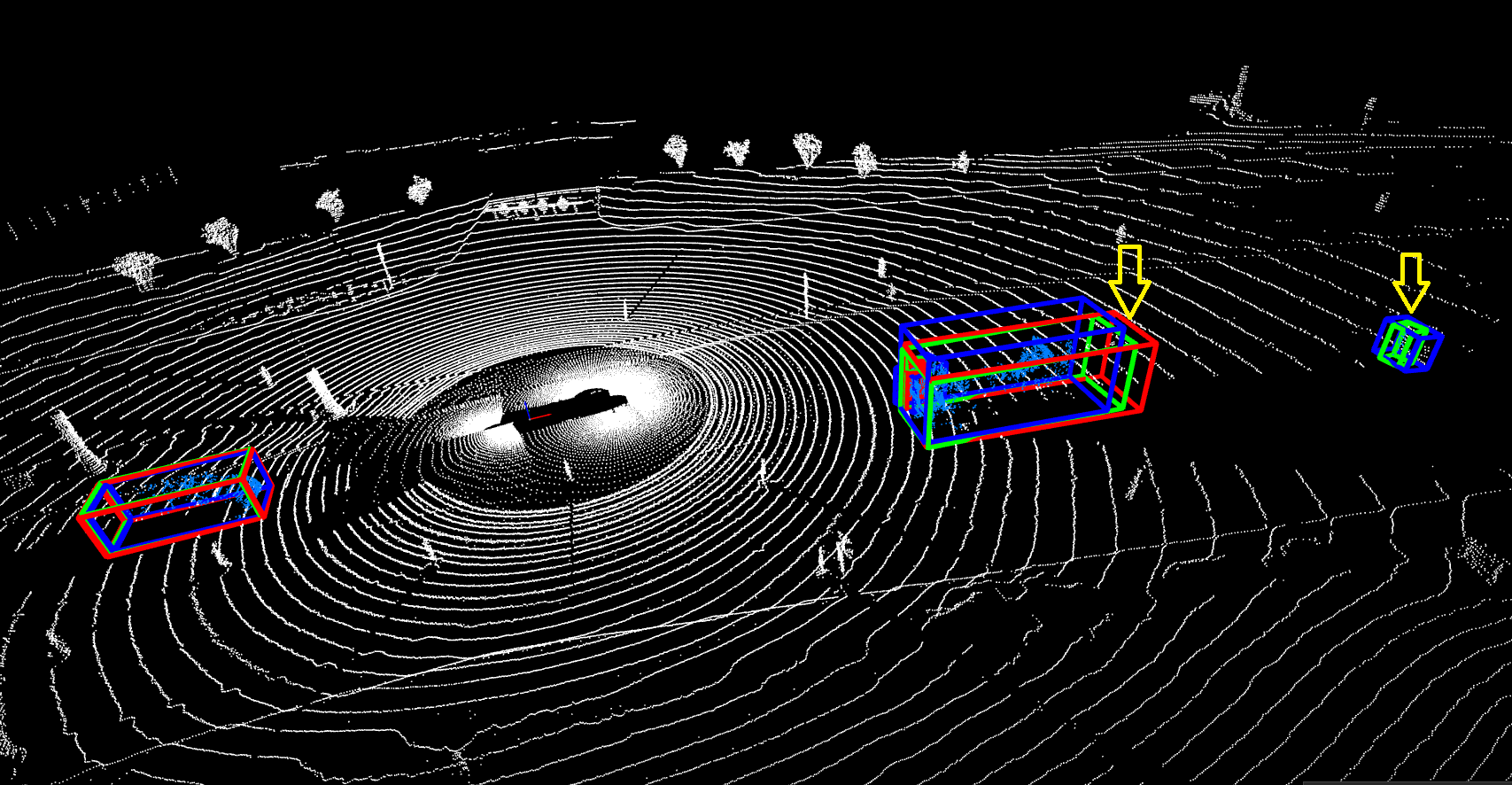}\label{fig:sub5}}
    \hfill
    \subfloat{\includegraphics[width=0.49\textwidth, height=5cm]{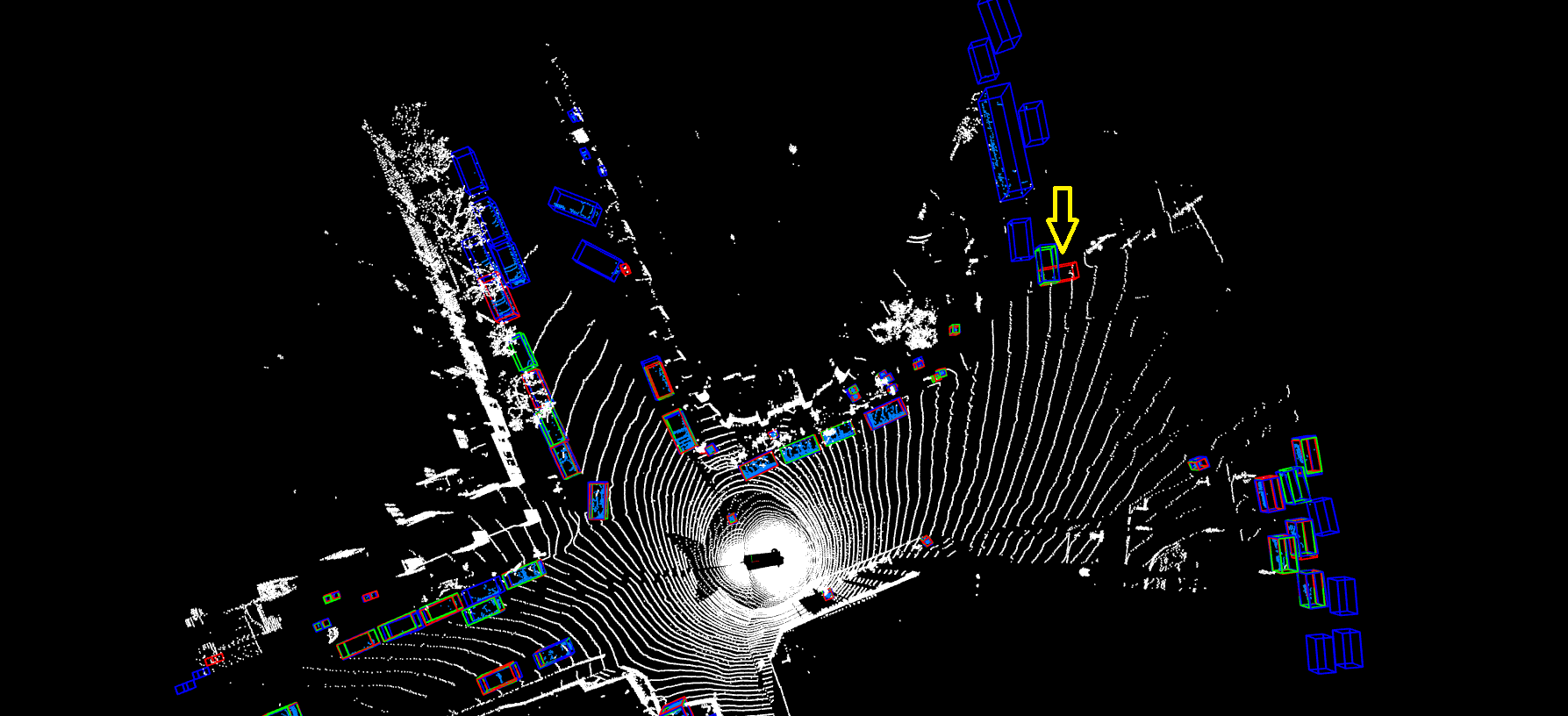}\label{fig:sub6}}

    \caption{A comparison of point cloud detection results between VDM and LION on the Waymo validation set. The blue, green, and red boxes represent human annotations, VDM predictions, and LION predictions, respectively. Yellow arrows on the figures highlight the differences in detection results. In the point cloud visualization process, the score threshold is set to 0.3, and the non-maximum suppression (NMS) threshold is set to 0.1.}
    \label{fig:main}
\end{figure*}

\noindent \textbf{Qualitative Visualization.}
To visually illustrate the advanced capabilities of our proposed model (VDM), we present a qualitative comparison of its detection performance against LION~\cite{liu2024lion} on the Waymo validation set. As depicted in Figure~\ref{fig:main}, the annotations marked by yellow arrows highlight several advantages of our method. Firstly, the bounding boxes predicted by VDM exhibit a higher degree of alignment with the ground truth annotations. Secondly, VDM is capable of correctly detecting objects (true positives) that are entirely missed by LION. Finally, our model demonstrates a lower false positive rate, indicating superior precision.

\noindent \textbf{Runtime Comparison.}
Table~\ref{runtime} compares the inference time of our VDM with LION~\cite{liu2024lion}. While VDM's voxel densification approach increases latency by processing more voxels, it offers a favorable accuracy-efficiency trade-off. On the nuScenes and ONCE datasets, latency increases by only 0.08s for improvements of 0.4 in NDS and 1.0 in mAP. For long-range detection on datasets like Waymo and Argoverse 2, the more substantial latency is justified by significant performance gains of 0.8 mAPH (L2) and 0.8 mAP, respectively.

\subsection{Ablation Study}

\noindent \textbf{Ablation Study on Voxel Densification.}
To assess the contribution of \textit{voxel densification} to detection performance, we conduct ablation experiments on the Waymo validation set using a fixed voxel size of [0.32, 0.32, 0.1875]. We evaluate the model both with and without the densification mechanism. As shown in Table \ref{ablity-diffusion}, introducing voxel expansion improves the VDM-OD model by approximately \textbf{1.3 mAPH (L2)}, underscoring its critical role in sequential detection frameworks. Moreover, on the nuScenes (Table \ref{nuscenes}) and Argoverse 2 (Table \ref{argov2}) datasets, we observe that under the \textit{Only-Densification} setting, the model achieves impressive detection accuracy, further validating the significant impact of pre-serialization voxel expansion.


\noindent \textbf{Ablation on Fine-grained Feature Aggregation.} To further evaluate the fine-grained feature aggregation capability of VDM, we conducted experiments by refining the voxel resolution. Specifically, we reduced the voxel size along the \textit{x} and \textit{y} axes to one-fourth of the original size, while keeping the resolution along the \textit{z}-axis unchanged, with \textit{voxel expansion} still enabled.

As reported in Table~\ref{frain-comparsion}, on the ONCE dataset, the \textit{Base} model (VDM-OD) exhibits inferior performance across all three categories when only voxel expansion is applied. With the integration of fine-grained feature aggregation (\textit{Fine}), the model achieves a \textbf{+1.5 mAP} gain, demonstrating the overall effectiveness of finer voxel partitioning.

To delve deeper into these improvements, we analyze the performance per category. It is worth noting that while the simplified VDM-OD (densification-only) variant achieves impressive performance on rigid, large-scale objects such as Vehicles (e.g., 70.5 mAP on nuScenes test set), the full VDM framework demonstrates superior capability in detecting smaller, non-rigid objects. For instance, on the \textbf{ONCE validation set}, the full VDM outperforms VDM-OD by a significant margin on the \textit{Pedestrian} category (\textbf{+3.2 mAP} in Table~\ref{once}). We attribute this phenomenon to the inherent geometric differences between categories. Large rigid objects (Vehicles) benefit primarily from increased voxel density to recover structural completeness, which VDM-OD provides efficiently. In contrast, small and sparse objects (Pedestrians) possess fewer points and more complex local geometries. The fine-grained aggregation mechanism in the full VDM is essential for capturing these subtle local contexts, which might be overlooked by simple densification. Therefore, although VDM-OD offers a strong baseline for vehicles, the full VDM provides a more balanced and robust solution, particularly for vulnerable road users (VRUs) critical to autonomous driving safety.


\section{Conclusion}

In this paper, we propose the \textbf{Voxel Densification Module (VDM)}, a novel pre-serialization stem designed to bridge the spatial-serial gap in existing 3D detectors. By leveraging sparse 3D convolutions, VDM explicitly \textit{expands} the foreground voxel set to address the sparsity limitation inherent in serialized models, while simultaneously performing fine-grained spatial aggregation to enrich local context. We validate the universality of our approach by integrating VDM into both SSM-based (LION) and Transformer-based (DSVT) architectures. Extensive experiments across four large-scale benchmarks demonstrate that VDM consistently boosts detection performance. Our analysis reveals that while \textit{pre-serialization densification} serves as the primary driver for mitigating feature sparsity, the fine-grained aggregation mechanism provides specific gains in scenarios requiring richer local geometric details. These results confirm that explicit voxel densification is a generic and effective strategy for advancing point cloud object detection.

\bibliographystyle{IEEEtran} 
\bibliography{IEEEtranBST/IEEEexample} 

\end{document}